\documentclass[10pt,twocolumn,letterpaper]{article}

 \usepackage[pagenumbers]{cvpr} % To force page numbers, e.g. for an arXiv version

\usepackage[dvipsnames]{xcolor}

\definecolor{cvprblue}{rgb}{0.21,0.49,0.74}
\usepackage[pagebackref,breaklinks,colorlinks,citecolor=cvprblue]{hyperref}

\def\paperID{13487} % *** Enter the Paper ID here
\def\confName{CVPR}
\def\confYear{2024}

\title{ACC-ViT : Atrous Convolution's Comeback in Vision Transformers}

\author{Nabil Ibtehaz \footnote{Work done as an intern at Futurewei Technologies Inc.}\\
Purdue University\\
West Lafayette, IN, United States\\
{\tt\small nibtehaz@purdue.edu}
\and
Ning Yan\\
Futurewei Technologies Inc.\\
Santa Clara, CA, United States\\
{\tt\small nyan@futurewei.com}
\and
Masood Mortazavi\\
Futurewei Technologies Inc.\\
Santa Clara, CA, United States\\
{\tt\small masood.mortazavi@futurewei.com}
\and 
Daisuke Kihara\\
Purdue University\\
West Lafayette, IN, United States\\
{\tt\small dkihara@purdue.edu}
}

\begin{document}
\maketitle
\begin{abstract}

Transformers have elevated to the state-of-the-art vision architectures through innovations in attention mechanism inspired from visual perception. At present two classes of attentions prevail in vision transformers, regional and sparse attention. The former bounds the pixel interactions within a region; the latter spreads them across sparse grids. The opposing natures of them have resulted in a dilemma between either preserving hierarchical relation or attaining a global context. In this work, taking inspiration from atrous convolution, we introduce Atrous Attention, a fusion of regional and sparse attention, which can adaptively consolidate both local and global information, while maintaining hierarchical relations. As a further tribute to atrous convolution, we redesign the ubiquitous inverted residual convolution blocks with atrous convolution. Finally, we propose a generalized, hybrid vision transformer backbone, named ACC-ViT, following conventional practices for standard vision tasks. Our tiny version model achieves $\sim 84 \%$ accuracy on ImageNet-1K, with less than $28.5$ million parameters, which is $0.42\%$ improvement over state-of-the-art MaxViT while having $8.4\%$ less parameters. In addition, we have investigated the efficacy of ACC-ViT backbone under different evaluation settings, such as finetuning, linear probing, and zero-shot learning on tasks involving medical image analysis, object detection, and language-image contrastive learning. ACC-ViT is therefore a strong vision backbone, which is also competitive in mobile-scale versions, ideal for niche applications with small datasets.

{\let\thefootnote\relax\footnote{{*Work done as an intern at Futurewei Technologies Inc.}}}

%The source code and trained models will be available upon publication.

\end{abstract}

\section{Introduction}
\label{sec:intro}

The field of computer vision has been undergoing several dramatic paradigm shifts and phenomenal innovations in recent years. Ever since the triumph of AlexNet in ImageNet competition \cite{Krizhevsky2012ImageNetNetworks}, convolutional neural networks (CNN) had been dominating and steadily progressing the different subdomains of vision \cite{He2016DeepRecognition,Ronneberger2015U-Net:Segmentation,He2017MaskR-CNN,Tan2019EfficientNet:Networks}. However, several radical breakthroughs in the last couple of years have substantially refined the existing problem-solving protocols \cite{Song2019GenerativeDistribution,Radford2021LearningSupervision,Dhariwal2021DiffusionSynthesis,Chen2022AVideos,Kirillov2023SegmentAnything}. One of the most influential discoveries has been the adaptation transformer architectures \cite{Vaswani2017AttentionNeed} from natural language processing (NLP) to vision \cite{Dosovitskiy2021AnScale}. The original Vision Transformer (ViT) explored the feasibility of leveraging the seemingly infinite scalability \cite{Fedus2021SwitchSparsity} of text transformers for processing images. Analogous to texts, the input image was broken into non-overlapping patches (i.e., words), embeddings for patches were computed (i.e., word embeddings) and the patches were treated as a sequence (i.e., sentence). Hence, following the practices of NLP \cite{Devlin2019BERT:Understanding}, the pipeline became suitable for vision transformers.

Although the feasibility of using transformers for analyzing images was established, earlier approaches lacked sufficient inductive bias and were devoid of any vision-specific adjustments \cite{Touvron2020TrainingAttention}. Hence, they fell short to state-of-the-art CNN models \cite{Graham2021LeViT:Inference}. One particular aspect those former ViTs overlooked was local patterns, which usually carry strong contextual information \cite{Li2021LocalViT:Transformers}. This led to the development of windowed attention, proposed by Swin transformer \cite{Liu2021SwinWindows}, the first truly competent vision transformer model. Instead of applying global attention over disjoint image patches, local  attention is computed between pixels within a window, analogous to convolution operation computes. Thus, this concept went through further innovations and eventually deviated towards two directions, namely, regional and sparse attention. In regional attention, windows of different sizes are considered to compute local attention at different scales \cite{Chu2021Twins:Transformers,Chen2021RegionViT:Transformers,Yang2021FocalTransformers}, whereas sparse attention computes a simplified global attention among pixels spread across a sparse grid \cite{Tu2022MaxViT:Transformer,Wang2021CrossFormer:Attention,Wang2023CrossFormer++:Attention}. Regional attention contains a sense of hierarchy \cite{Chen2021RegionViT:Transformers}, but the global view is harshly limited due to only considering a couple of regions to account for computational complexity \cite{Yang2021FocalTransformers}. On the contrary, sparse attention can compute a better approximate global context with reasonable computational cost, but sacrifices nested interaction across hierarchies \cite{Wang2021CrossFormer:Attention}. Although hierarchical information is immensely valuable \cite{Yang2022FocalNetworks}, owing to the access to a richer global context, sparse attention is used in state-of-the-art models and a relatively small-sized window attention can compensate for the limited local interaction \cite{Tu2022MaxViT:Transformer}.

%Regional attention despite being the logical choice, falls short in performance for missing out global context. Sparse context on the other hand, achieves state-of-the-art performance but lacks intuition. Therefore, it is motivating to develop something in between, injecting sparsity into regional attention or similarly, bounding sparse attention by regions. Such an attention mechanism likely has the potential to balance the best of both worlds, retaining hierarchical information while also approximating a global context. This has been the central objective of this work, making a fusion of sparse and regional attention.
\begin{figure}[h]
    \centering
    \includegraphics[width=\columnwidth]{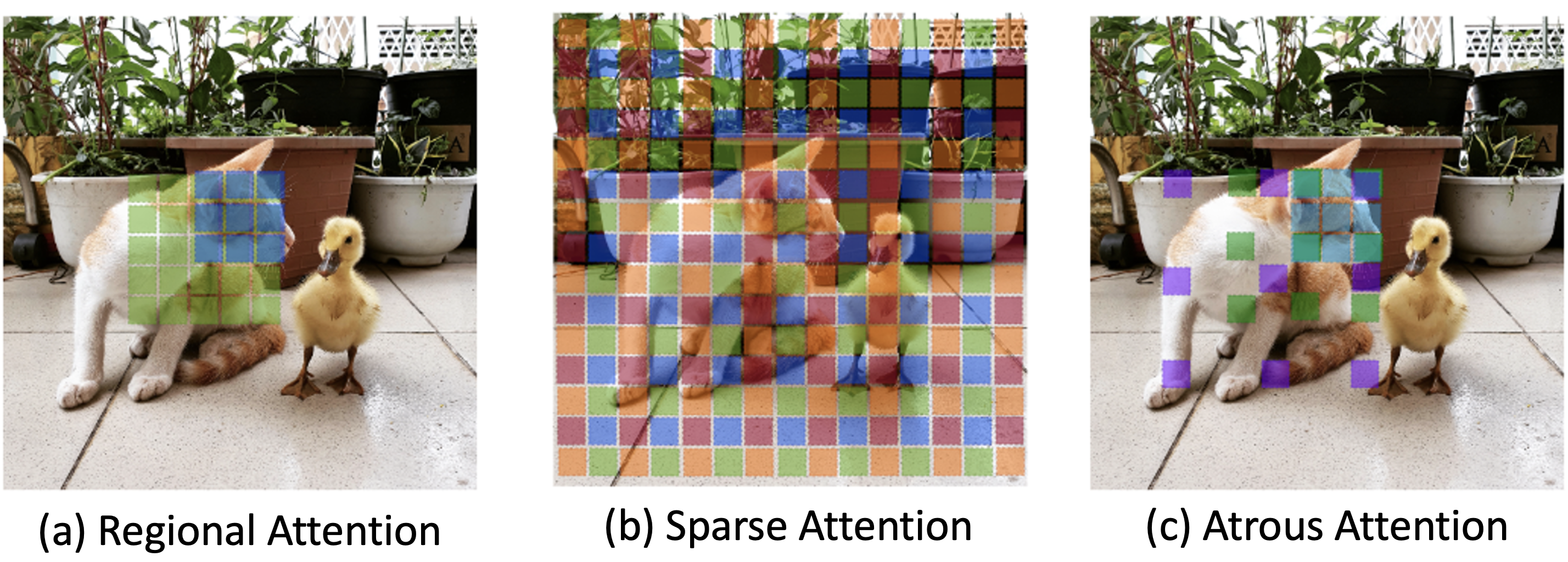}
    \caption{Simplified illustrations of different types of windowed attention mechanisms.}
    \label{fig:attn}
\end{figure}

In Fig. \ref{fig:attn}, we have presented simplified examples of regional and sparse attention. In regional attention (Fig. \ref{fig:attn}a), the cat is identified by two patches. The blue and the green patches inspect the head and most of the cat's body, respectively. By combining the information hierarchy, we can deduce that it is a cat. However, the legs and the tail have been missed by regional attention due to the high cost of computing attention over large regions. In the case of sparse attention (Fig. \ref{fig:attn}b), we can observe that the entire image is being analyzed by sets of grids marked by four different colors. Although seemingly, we can compute attention over the entire cat's body, there hardly exists information hierarchy and a good deal of irrelevant pixels are also considered. For instance, computing attention over the duck pixels probably does not contribute much to classifying the cat.

Based on the above intuitive analysis, we can make the following two observations. Firstly, it is preferable to cover more regions, with reasonable computational expense. Secondly, while global information is beneficial, we may probably benefit by limiting our receptive field up to a certain extent. Combining these two observations, we can thus induce sparsity in inspected regions or make the grids bounded. These reduce to dilated regions, which have been analyzed in atrous convolution \cite{Chen2016DeepLab:CRFs}. Therefore, taking inspiration from atrous convolution, we propose Atrous Attention. We can consider multiple windows, dilated at varying degrees, to compute attention and then combine them together. This solves the two issues concurrently, maintaining object hierarchy and capturing global context. For example, in Fig. \ref{fig:attn}c, it can be observed that 3 smaller patches with different levels of dilation are capable of covering all of the cat.

We propose a novel attention mechanism for vision transformers, and we further design a hybrid vision transformer, ACC-ViT, based on that attention mechanism. We turn back to almost obscured atrous (or dilated) convolution, in the vision transformer era, and discover that the attributes of atrous convolution are quite beneficial for vision transformers. We thus design both our attention mechanism and convolution blocks based on the atrous convolution. In addition, taking inspiration from atrous spatial pyramid pooling (ASPP) \cite{Chen2016DeepLab:CRFs}, we experiment with parallel design, deviating from the existing style of stacking different types of transformer layers \cite{Tu2022MaxViT:Transformer}. ACC-ViT outperforms state-of-the-art ViT models, such as, MaxViT \cite{Tu2022MaxViT:Transformer} and MOAT \cite{Yang2022MOAT:Models} on ImageNet-1K (Fig. \ref{fig:pareto}). In addition, ACC-ViT has been evaluated on a diverse set of downstream tasks, including finetuning, linear probing, and zero-shot learning. Moreover, when scaled down, ACC-ViT still maintains its competency, all of which make ACC-ViT suitable for diverse applications across different domains.

\section{Related Works}
\label{sec:related_works}

\subsection{CNN vs ViT vs Hybrid models}

The historic AlexNet model \cite{Krizhevsky2012ImageNetNetworks}, a convolutional network, dominated the ImageNet competition and brought a revolution in computer vision. Over the years numerous CNN models were developed to perform diverse vision tasks and achieved unprecedented levels of performance \cite{Szegedy2015GoingConvolutions,
Ronneberger2015U-Net:Segmentation,Oh2015Action-ConditionalGames,He2016DeepRecognition,Zhao2018ThePixels,Li2021COMISR:Super-Resolution,Wang2021RichVideos,Chen2021ProxIQA:Compression}, making them the \textit{de facto} standard. Concurrently, natural language processing (NLP) was disrupted by transformers \cite{Vaswani2017AttentionNeed} due to their seemingly infinite scalability \cite{Fedus2021SwitchSparsity}. As a result, transformers kept on accumulating attention from a myriad of communities and were adopted for various tasks and data modalities \cite{Lee2019SetNetworks,Yun2019GraphNetworks}. The rising interest for transformers eventually led to the successful use of transformers in vision \cite{Dosovitskiy2021AnScale}, and also managed to outperform competent convolutional models \cite{Liu2021SwinWindows}. The lack of sufficient perceptive inductive bias in transformers were compensated by using convolution operation in transformers \cite{Xiao2021EarlyBetter}, which led to a new class of vision architectures, namely hybrid models \cite{Dai2021CoAtNet:Sizes}. The debate between transformers vs CNNs still continued \cite{Zhou2021ConvNetsTransferable} and it was actually discovered that transformer-like architecture design and training protocol substantially improves CNN models \cite{Liu2022A2020s,Woo2023ConvNeXtAutoencoders}. Nevertheless, hybrid models are still the overall well-rounded models and fusing the benefits of both the architectures, they are the state-of-the-art models \cite{Tu2022MaxViT:Transformer,Yang2022MOAT:Models}.

\begin{figure}[h]
    \centering
\includegraphics[width=\columnwidth]{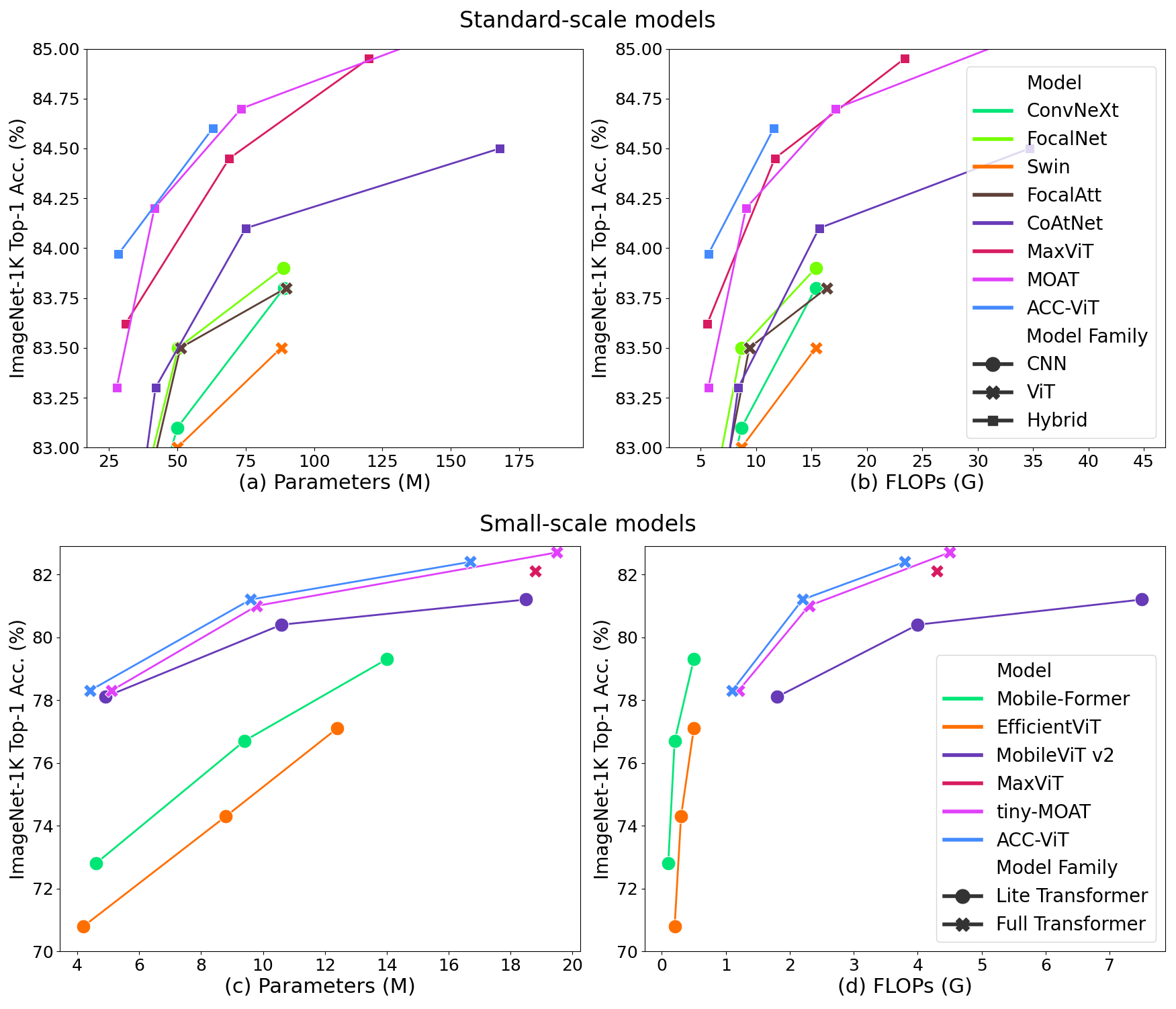}
    \caption{ACC-ViT performs competitively against both standard and small-scale models on ImageNet-1K.}
    \label{fig:pareto}
\end{figure}

\subsection{Attention mechanisms for ViTs}

The original vision transformer was based on solely global attention across disjoint image patches \cite{Dosovitskiy2021AnScale}, and locality was later analyzed \cite{Li2021LocalViT:Transformers}. Swin transformer \cite{Liu2021SwinWindows}, brought a breakthrough and introduced windowed attention. Successive developments had undergone, investigating the best possible approach to compute windowed attention. Focal attention \cite{Yang2021FocalTransformers}  considered multiple windows of different size to preserve the sense of hierarchy. Twins \cite{Chu2021Twins:Transformers} architecture leveraged the local-global interaction through locally-grouped and globally-subsampled attention. RegionViT \cite{Chen2021RegionViT:Transformers} implemented an efficient attention to inject global context into the local information. While these methods focused more on the regional aspect of images, a different class of attention based on sparsity was proposed to capture the global view better. CrossFormer \cite{Wang2021CrossFormer:Attention}, CrossFormer++ \cite{Wang2023CrossFormer++:Attention}, and the current state-of-the-art MaxViT \cite{Tu2022MaxViT:Transformer} all uses gapped, sparsified attention mechanism in a grid-like manner. Apart from these conventional forms of attention, in recent years some investigations have been conducted making the attention computation adaptive, for instance, through instance-dependency \cite{Wei2023Sparsifiner:Transformers} and deformation \cite{Xia2022VisionAttention}.

\subsection{Atrous Convolution}

The concept of atrous or dilated convolution from traditional signal and image processing \cite{Holschneider1990ATransform} was properly introduced in computer vision by deeplab \cite{Chen2016DeepLab:CRFs}, which quickly gained popularity. By ignoring some consecutive neighboring pixels, atrous convolution increases the receptive field in deep networks \cite{Luo2017UnderstandingNetworks}. Moreover, Atrous spatial pyramid pooling can extract hierarchical features \cite{Chen2017RethinkingSegmentation}, which can also be employed adaptively based on conditions \cite{Qiao2020DetectoRS:Convolution}. However, in the ViT era, the use of atrous convolution has severely declined, limited only up to patch computation \cite{Huang2022AtrousInpainting}.

\section{Methods}
\label{sec:methods}

\begin{figure*}
    \centering
    \includegraphics[width=0.9\textwidth]{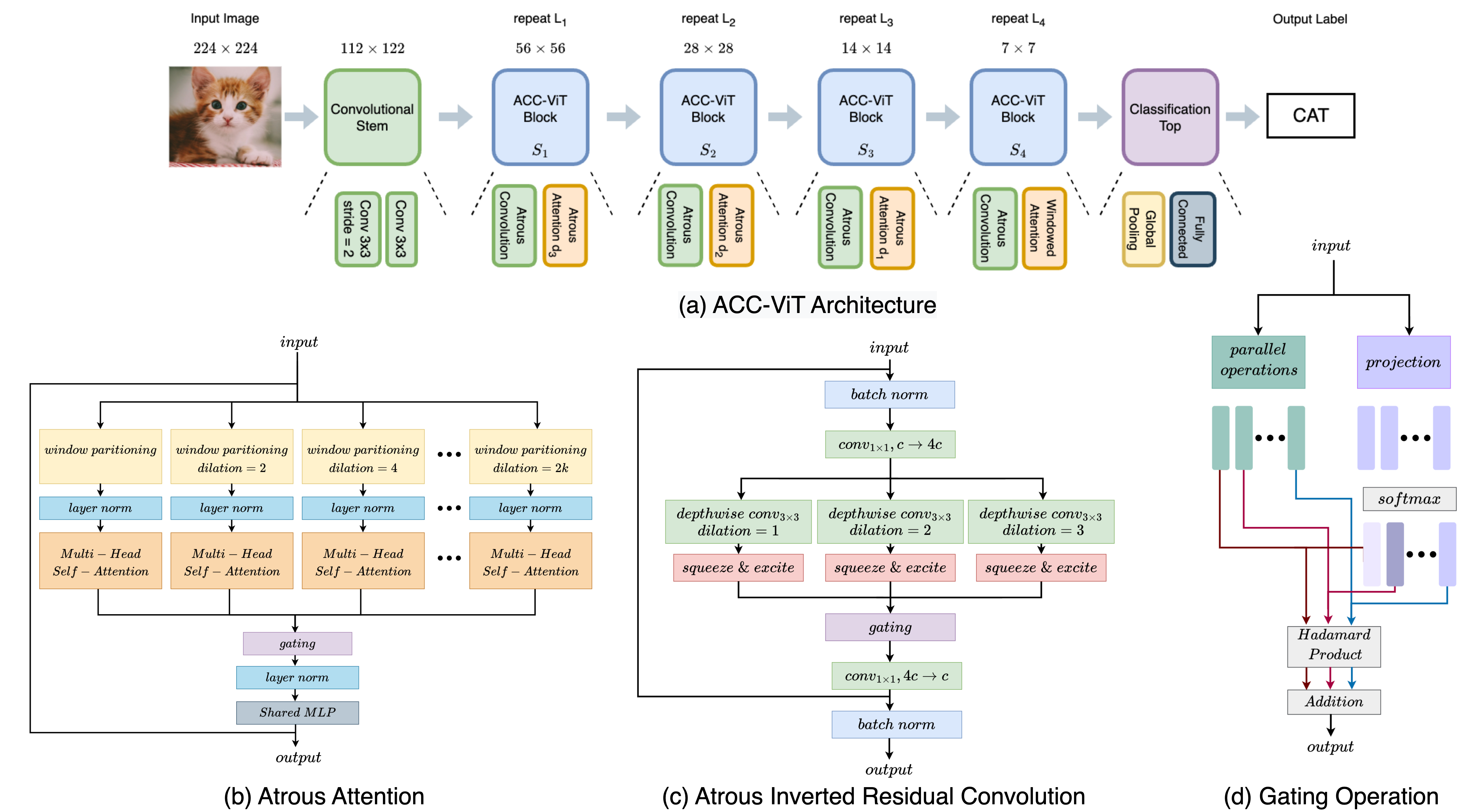}
    \caption{ACC-ViT model architecture and its components.}
    \label{fig:mdl-dia}
\end{figure*}

\subsection{Atrous Attention}

In this paper, we propose a new form of attention mechanism for vision transformers, attempting to perform a fusion between regional and sparse attention. This enables us to inspect over a global context with reasonable computational complexity, while also retaining the hierarchical information scattered throughout the image. We perform this by emulating the way atrous convolution is computed and name our attention mechanism Atrous Attention.

Atrous Attention is inspired from atrous convolution \cite{Chen2016DeepLab:CRFs}, which drops some rows and columns from an image to effectively increase the receptive field, without additional parameters. We mimic a similar strategy during window partitioning of the featuremaps. For a given input featuremap $x \in \mathcal{R}^{c,h,w}$, where $c,h,w$ refers to the number of channels, height, and width, respectively, we can compute windows at different levels of dilation as,
\begin{equation}
    x_{atr-k}[c,i,j] = x[c,i+2k-1,j+2k-1]
\end{equation}
For example, 
\begin{equation}
    x_{atr-1}[c,i,j] = x[c,i+1,j+1]
\end{equation}
\begin{equation}
    x_{atr-2}[c,i,j] = x[c,i+3,j+3]
\end{equation}
\begin{equation}
    x_{atr-3}[c,i,j] = x[c,i+7,j+7]
\end{equation}
Here, we considered discarding $2^k-1$ number of neighboring pixels during the window partitioning, which is treated as $2^k$ dilation rate on conventional convolutional layers \cite{Paszke2019PyTorch:Library}. We use this strategy because such dilations can be efficiently computed on GPU using einops operation \cite{Rogozhnikov2022Einops:Notation}.

Similar to atrous convolution, all the values can be captured by shifting or sliding the dilated patterns of $x_{atr-k}$ across row and column $k^2$ times. We apply windowed multi-head self-attention with relative positional embedding \cite{Shaw2018Self-AttentionRepresentations}, on each of the windows computed at different levels of dilation ($x_{atr-k}$) along with the undilated input ($x$). 
\begin{equation}
    y_0 = W-MHSA(LN(x)) + x
\end{equation}
\begin{equation}
    y_k = W-MHSA(LN(x_{atr-k})) + x_{atr-k}
\end{equation}
Here, $y_k,k>0$ denotes the output of Atrous Attention at different levels of dilation $x_{atr-k}$, $y_0$ being the output of undilated input $x$. $W-MHSA$ and $LN$ corresponds to windowed multi-head self-attention \cite{Liu2021SwinWindows} and layer normalization \cite{Ba2016LayerNormalization}, respectively. For example,
\begin{equation}
    y_1 = W-MHSA(LN(x_{atr-1})) + x_{atr-1}
\end{equation}
\begin{equation}
    y_2 = W-MHSA(LN(x_{atr-2})) + x_{atr-2}
\end{equation}
%Atrous Attention is therefore a fusion of regional and sparse attention. The attentions are computed over different sizes of windowed-regions, preserving the levels of hierarchy. Moreover, during window partition, sparsity has been leveraged, enabling the extraction of features from substantially larger contexts, minimizing the number of parameters.

\subsection{Gating Operation for Adaptive Fusion of Different Branches}
\label{sec:gating}
In our proposed attention, we have attended to an input image and/or featuremap from multiple levels of hierarchy. Fusing these computed features through averaging or concatenating misses fine details or increases computational cost, respectively \cite{Ibtehaz2023ACC-UNet:Forthe2020s}. Recent works have demonstrated that proper aggregation of hierarchical context can substantially improve the visual representation of vision models \cite{Yang2022FocalNetworks}. Therefore, we have developed a simple, lightweight, and adaptive gating operation to merge the features.

Based on the input featuremap, $x$, $g \in [0 \sim 1]$ weight factors are computed for each output branch in the form of a gating operation. This is performed by non-linearly computing a weight factor for each individual spatial and filter values of the input featuremap, normalizing them across the output branches through softmax. The gating operation, denoted as $g$, can be learned through back-propagation, which introduces learnability and adaptability in contrast to static aggregation. We hypothesize that based on the input, the gating operation understands which portions of the different computed features should be emphasized or discarded.

Formally, an input tensor $x \in \mathcal{R}^{c,h,w}$ is processed by $k$ independent non-linear operations ($ops_i$), such that the dimensions are unchanged, i.e., $y_i = ops_i(x), \text{ where }y_i \in \mathcal{R}^{c,h,w},\text{ for } i=1,\cdots,k$. The gating function $g \in \mathcal{R}^{k,c,h,w}$ is computed as follows, through Softmax across the $k$ dimensions:
\begin{equation}
    g = Softmax_k(Activation(Linear_{chw \rightarrow kchw}(x)))
\end{equation}
The fused featuremap $y_{fused}$, adaptively amalgamated from the different branches, can thus be computed from a Hadamard product or elementwise multiplication between the different feature branches with the appropriate gating function, across channels and spatial dimensions.
\begin{equation}
    y_{fused} = \sum_{i=1}^{k} g_i \odot y_i
\end{equation}
\begin{table*}[h]
\centering
\caption{Different Configurations of ACC-ViT model.}
\label{tbl:arch}
\begin{tabular}{l|c|ccc|ccc}
\hline
\textbf{Stage} & \textbf{Size} & \textbf{tiny} & \textbf{small} & \textbf{base} & \textbf{nano} & \textbf{pico} & \textbf{femto} \\ \hline
\textbf{S0: Stem}     & 1/2  & B=2, C=64  & B=2, C=64  & B=2, C=64   & B=2, C=64  & B=2, C=48  & B=2, C=32  \\
\textbf{S1: Block 1}  & 1/4  & B=2, C=64  & B=2, C=96  & B=4, C=96   & B=1, C=64  & B=1, C=48  & B=1, C=32  \\
\textbf{S2: Block 2}  & 1/8  & B=3, C=128 & B=3, C=192 & B=6, C=192  & B=2, C=128 & B=2, C=96  & B=2, C=64  \\
\textbf{S3: Block 3}  & 1/16 & B=6, C=256 & B=6, C=384 & B=14, C=384 & B=4, C=256 & B=4, C=192 & B=4, C=128 \\
\textbf{S4: Block 4}  & 1/32 & B=2, C=512 & B=2, C=768 & B=2, C=768  & B=1, C=512 & B=1, C=384 & B=1, C=256 \\
\textbf{\#params (M)} &      & 28.367     & 62.886     & 103.576     & 16.649     & 9.55       & 4.4        \\
\textbf{FLOPs (G)}    &      & 5.694      & 11.59      & 22.316      & 3.812      & 2.217      & 1.049     \\ \hline
\end{tabular}
\end{table*}

\subsection{Shared MLP Layer across Parallel Attentions}
In conventional transformer layers, an MLP (multi layer perceptron) layer follows multi-head self-attention layers, which combines the information across the different heads. In our formulation, instead of using MLP layers after each attention operation, we apply a shared MLP layer on top of the fused attention map, obtained from the gating operation. This not only reduces the computational complexity, but also makes it easier for the model to learn. We conjecture that when merging the featuremaps mixed through different MLPs, the model has to reconcile a higher degree of variation, compared to fusing the activations from the individual attention heads, focused on specific aspects (Section \ref{sec:abla}). The output of Atrous Attention thus becomes, 
\begin{equation}
    y_{out} = MLP(LN(y_{fused})) + y_{fused}
\end{equation}
\subsection{Parallel Atrous Inverted Residual Convolution}

Inverted residual convolution is the \textit{de facto} choice of convolution in vision transformers \cite{Tu2022MaxViT:Transformer,Yang2022MOAT:Models} due to their efficiency \cite{Howard2017MobileNets:Applications} and high accuracy \cite{Tan2019EfficientNet:Networks}. As a hybrid model, our model also benefits from convolutional layers based on the inverted residual or mobile convolution. However, in order to exploit sparsity and hierarchy throughout the model, we replace the intermediate depthwise separable convolution with 3 parallel atrous, depthwise separable convolution, with dilation rates of 1 (no dilation), 2, and 3, respectively. The outputs of the three convolutional operations are merged using the same gating operation, described in Section \ref{sec:gating}. Thus, formally our proposed Atrous Inverted Residual Convolution block can be presented as follows:
\begin{equation}
    y = x + SE(Conv_{1 \times 1}^{c\times 0.25}( G (DConv_{3 \times 3}^{dil=1,2,3} (Conv_{1 \times 1}^{c\times 4}(x)))))
\end{equation}
Here, $SE$ and $G$ refers to squeeze-and-excitation block \cite{Hu2018Squeeze-and-ExcitationNetworks} and gating operation, respectively. The subscripts of $Conv$ and $DConv$ denote the kernel size, and the superscripts refer to the expansion ratio or dilation rates, where applicable.

\subsection{ACC-ViT}

Using the proposed attention and convolution blocks, we have designed ACC-ViT, a hybrid, hierarchical vision transformer architecture. Following conventions, the model comprises a convolution stem at the input level, 4 ACC-ViT blocks, and a classification top \cite{Tu2022MaxViT:Transformer}. The stem downsamples the input image, making the computation of self-attention feasible. The stem is followed by 4 ACC-ViT blocks, which are built by stacking Atrous Convolution and Attention layers, and the images are downsampled after each block. Since the image size gets reduced, we use 3,2,1 and 0 (no dilation) levels of Atrous Attention in the 4 blocks, respectively. Finally, the classifier top is based on global average pooling and fully connected layers, as used in contemporary architecture \cite{Tu2022MaxViT:Transformer,Liu2021SwinWindows,Liu2022A2020s}. A diagram of our proposed architecture is presented in Fig. \ref{fig:mdl-dia}.

Following the conventional practices, we have designed several variants of ACC-ViT, having different levels of complexity, the configurations are presented in Table \ref{tbl:arch}. 

\section{Experiments and Results}
\label{sec:results}

We have implemented the ACC-ViT model and trained it on the ImageNet-1K database \cite{Deng2009ImageNet:Database} using the Torchvision library \cite{TorchVisionmaintainersandcontributors2016TorchVision:Library}. The experiments were run on a server computer having 2x Xeon 8268 24core @2.9GHz CPUs and 16x 32GB Nvidia V100 GPUs. Additionally, we have validated the efficacy of ACC-ViT on various vision tasks, under different settings, namely, finetuning, linear probing, and zero-shot learning. We conducted those experiments only on the tiny-size models due to computational limitations. The performance of ACC-ViT was compared against the most widely adopted ViT baseline, Swin Transformer along with ConvNeXt and MaxViT, which are two state-of-the-art CNN and ViT models, respectively, available through Torchvision. All our experiments are based on open-source tools, please refer to the Appendix for more details.

\subsection{ImageNet Image Classification}

We have trained the different variants of ACC-ViT on the standard image classification task, using the ImageNet-1K dataset \cite{Deng2009ImageNet:Database}. The models were trained using a combination of the hyperparameter values recommended by Torchvision \cite{TorchVisionmaintainersandcontributors2016TorchVision:Library} and as adopted in recent works \cite{Yang2022FocalNetworks, Tu2022MaxViT:Transformer, Yang2022MOAT:Models}. Under the conventional supervised learning paradigm, the models were trained on $224 \times 224$ resolution images, without any extra data. The top-1 validation accuracy achieved by ACC-ViT is presented in Table \ref{tbl:res_imgnet}, along with a comparison against popular state-of-the-art architectures.

\begin{table}[h]
\centering
\caption{Top-1 classification accuracy on ImageNet-1K dataset. The models are divided into 3 categories based on model size.}
\label{tbl:res_imgnet}
\begin{tabular}{cccc}
\hline
Model                                         & \#Params                   & FLOPs & \begin{tabular}[c]{@{}c@{}}IN-1K\\ Val Acc.\end{tabular} \\ \hline
ResNet-50-SB \cite{Wightman2021ResNetTimm}   & 25M    & 4.1G  & 79.8  \\
ConvNeXt-T \cite{Liu2022A2020s}              & 29M    & 4.5G  & 82.1  \\
FocalNet-T \cite{Yang2022FocalNetworks}      & 28.6M  & 4.5G  & 82.3  \\ 
PVT-Small \cite{Wang2021PyramidConvolutions} & 24.5M  & 3.8G  & 79.8  \\
Swin-T \cite{Liu2021SwinWindows}             & 28M    & 4.5G  & 81.3  \\
PoolFormer-S36 \cite{Yu2021MetaFormerVision} & 31M    & 5.0G  & 81.4  \\
FocalAtt-T  \cite{Yang2021FocalTransformers} & 28.9M  & 4.9G  & 82.2  \\ 
CoAtNet-0 \cite{Dai2021CoAtNet:Sizes}        & 25M    & 4.2G  & 81.6  \\
MaxViT-T \cite{Tu2022MaxViT:Transformer}     & 31M    & 5.6G  & 83.62 \\
MOAT-0 \cite{Yang2022MOAT:Models}            & 27.8M  & 5.7G  & 83.3  \\
ACC-ViT-T                                    & 28.4M  & 5.7G  & \textbf{83.97} \\ \hline
ResNet-101-SB \cite{Wightman2021ResNetTimm}  & 45M  & 7.9G  & 81.3  \\
ConvNeXt-S \cite{Liu2022A2020s}              & 50M    & 8.7G  & 83.1  \\
FocalNet-S \cite{Yang2022FocalNetworks}      & 50.3M  & 8.7G  & 83.5  \\ 
PVT-Medium \cite{Wang2021PyramidConvolutions} & 44.2M & 6.7G  & 81.2 \\
Swin-S \cite{Liu2021SwinWindows}             & 50M    & 8,7G  & 83.0  \\
PoolFormer-M36 \cite{Yu2021MetaFormerVision} & 56M    & 8.8G  & 82.1  \\
FocalAtt-S \cite{Yang2021FocalTransformers}  & 51.1M  & 9.4G  & 83.5  \\ 
CoAtNet-1 \cite{Dai2021CoAtNet:Sizes}        & 42M    & 8.4G  & 83.3  \\
MaxViT-S \cite{Tu2022MaxViT:Transformer}     & 69M    & 11.7G & 84.45 \\
MOAT-1 \cite{Yang2022MOAT:Models}            & 41.6M  & 9.1G  & 84.2  \\
ACC-ViT-S                                    & 62.9M  & 11.6G & \textbf{84.6}  \\ \hline
ResNet-152-SB \cite{Wightman2021ResNetTimm}  & 60M    & 11.6G & 81.8  \\
ConvNeXt-B \cite{Liu2022A2020s}              & 89M    & 15.4G & 83.8  \\
FocalNet-B \cite{Yang2022FocalNetworks}      & 88.7M  & 15.4G & 83.9  \\ 
PVT-Large \cite{Wang2021PyramidConvolutions} & 61.4M  & 9.8G  & 81.7  \\
Swin-B \cite{Liu2021SwinWindows}             & 88M    & 15.4G & 83.5  \\
PoolFormer-M48 \cite{Yu2021MetaFormerVision} & 73M    & 11.6G & 82.5  \\
FocalAtt-B \cite{Yang2021FocalTransformers}  & 89.8M  & 16.4G & 83.8  \\ 
CoAtNet-2 \cite{Dai2021CoAtNet:Sizes}        & 75M    & 15.7G & 84.1  \\
CoAtNet-3 \cite{Dai2021CoAtNet:Sizes}        & 168M   & 34.7G & 84.5  \\
MaxViT-B \cite{Tu2022MaxViT:Transformer}     & 120M   & 23.4G & 84.95 \\
MOAT-2 \cite{Yang2022MOAT:Models}            & 73.4M  & 17.2G & 84.7  \\
MOAT-3  \cite{Yang2022MOAT:Models}           & 190M   & 44.9G & \textbf{85.3}  \\
%\st{ACC-ViT-B & 103.6M & 22.3G &  85} \\ 
\hline
\end{tabular}
\end{table}

We have compared ACC-ViT with top-performing models of different types, which include 3 CNNs (ResNet optimized by \cite{Wightman2021ResNetTimm}, ConvNeXt \cite{Liu2022A2020s}, FocalNet \cite{Yang2022FocalNetworks}), 4 ViTs (PVT \cite{Wang2021PyramidConvolutions}, Swin \cite{Liu2021SwinWindows}, PoolFormer \cite{Yu2021MetaFormerVision}, FocalAtt \cite{Yang2021FocalTransformers}) and 3 hybrid models (CoAtNet \cite{Dai2021CoAtNet:Sizes}, MaxViT \cite{Tu2022MaxViT:Transformer}, MOAT \cite{Yang2022MOAT:Models}), resulting in an elaborate comparison covering several novel ideas and architectures, proposed in recent time.

ACC-ViT outperforms the existing state-of-the-art MaxViT and MOAT models within similar ranges of parameters and FLOPs. In tiny and small models, using a similar amount of flops, ACC-ViT achieves $0.35\%$ and $0.15\%$ higher accuracy than MaxViT, despite having $9.15\%$ and $9.7\%$ less parameters. For all the other models, the performance of ACC-ViT is even more impressive. The tiny and small versions of ACC-ViT are more accurate than the small and base variants of the other models, respectively.

\begin{table*}[h]
\centering
\caption{Transfer learning experiment results on 3 medical image datasets.}

\label{tbl:res_imgnet}
\begin{tabular}{c|cccc|cccc|cccc}
\cline{2-13}
                                & \multicolumn{4}{c|}{HAM10000} & \multicolumn{4}{c|}{EyePACS} & \multicolumn{4}{c}{BUSI} \\ \cline{2-13} 
                                & Pr   & Re   & F1  & Acc  & Pr   & Re   & F1   & Acc  & Pr   & Re   & F1  & Acc  \\ \hline
\multicolumn{1}{c|}{Swin-T}     &  82.34   &  75.93   & 78.37    &   88.32   &    \textbf{67.37}  &  48.94    &  52.85  & 83.05   &   75.18   &  75.70    &     75.07 &     77.56  \\
\multicolumn{1}{c|}{ConvNeXt-T} &   83.69   &   76.84   &   79.48  &  89.57    &  66.82    &  52.30    &  56.20    &   83.90   &    79.75  &     \textbf{79.65} &  \textbf{79.42}   &   \textbf{82.05}   \\
\multicolumn{1}{c|}{MaxVit-T}   &  64.45    &  57.87    &  59.71   &   84.62   &  64.63    &   51.43   & 53.99     &   83.67   &   72.30   &    63.75  &  66.46   &    73.08  \\
\multicolumn{1}{c|}{ACC-ViT-T}  &  \textbf{90.02}    &  \textbf{84.26}    &  \textbf{86.77}    &  \textbf{92.06}    &   66.42   &   \textbf{57.00}   &   \textbf{60.12}   &   \textbf{85.07}   &  \textbf{80.10}    &   76.30   &   77.80  &   80.77   \\ \hline
\end{tabular}
\end{table*}

\begin{table}[h]
\centering
\caption{Object Detection and Instance Segmentation performance on the tiny size models as frozen feature extractors.}
\label{fig:mmdet}
\begin{tabular}{l|ccc|ccc}
\hline                           & AP    & AP$_{50}$ & AP$_{75}$ & AP$^{m}$ & AP$^{m}_{50}$ & AP$^{m}_{75}$  \\ \hline
\multicolumn{7}{c}{Resolution : $224 \times 224$} \\ \hline
MaxViT          & 0.2   & 0.357   & 0.201   & 0.192  & 0.329       & 0.195       \\
ACC-ViT                                 & \textbf{0.21} & \textbf{0.366}   & \textbf{0.206}   & \textbf{0.193}  & \textbf{0.338}       & \textbf{0.196}         \\ \hline
\multicolumn{7}{c}{Resolution : $448 \times 448$} \\ \hline
MaxViT                              &   0.267 & 0.472 & 0.27 & 0.262 & 0.446 & 0.269      \\
ACC-ViT                                   & \textbf{0.278} & \textbf{0.485} & \textbf{0.281} & \textbf{0.269} & \textbf{0.458} & \textbf{0.279}   \\ \hline
\end{tabular}
\end{table}

\begin{table}[h]
\centering
\caption{ImageNet-1K classification by small-scale models.} %and the best performance achieved in each category is marked by bold.}
\label{tbl:smol}
\begin{tabular}{cccc}
\hline
Model & \#Params & FLOPs & \begin{tabular}[c]{@{}c@{}}IN-1K\\ Val Acc.\end{tabular} \\ \hline
EfficientViT-M2 \cite{Liu2023EfficientViT:Attention} & 4.2M  & 0.2G  & 70.8 \\
Mobile-Former-96M \cite{Chen2021Mobile-Former:Transformer}  & 4.6M  & 0.1G & 72.8 \\
MobileViT v2-1.0 \cite{Mehta2022SeparableTransformers}  & 4.9M  & 1.8G & 78.1 \\
%ConvNeXt v2-F      & 5.2M  & 0.8G & 78.5 \\
tiny-MOAT-1 \cite{Yang2022MOAT:Models}       & 5.1M  & 1.2G & 78.3 \\
ACC-ViT F       & 4.4M      &  1.1G   &  \textbf{78.3}    \\ \hline
EfficientViT-M4 \cite{Liu2023EfficientViT:Attention} & 8.8M  & 0.3G  & 74.3 \\
Mobile-Former-214M \cite{Chen2021Mobile-Former:Transformer} & 9.4M  & 0.2G & 76.7 \\
MobileVit v2-1.5 \cite{Mehta2022SeparableTransformers}  & 10.6M & 4G   & 80.4 \\
%ConvNeXt v2-P      & 9.1M  & 1.4G & 80.3 \\
tiny-MOAT-2 \cite{Yang2022MOAT:Models}       & 9.8M  & 2.3G & 81   \\
ACC-ViT P      &   9.6M    &   2.2G   &    \textbf{81.2}  \\ \hline
EfficientViT-M5 \cite{Liu2023EfficientViT:Attention} & 12.4M  & 0.5G  & 77.1 \\
Mobile-Former-508M \cite{Chen2021Mobile-Former:Transformer} & 14.0M & 0.5G & 79.3 \\
MobileVit v2-2.0 \cite{Mehta2022SeparableTransformers}  & 18.5M & 7.5G & 81.2 \\
%ConvNeXt v2-N      & 15.6M & 2.5G & 81.9 \\
tiny-MOAT-3 \cite{Yang2022MOAT:Models}       & 19.5M & 4.5G & \textbf{82.7} \\
MaxViT N        &  18.8M     &  4.3G    &   82.1  \\
ACC-ViT N       &    16.7M   &  3.8G    &   82.4   \\ \hline
\end{tabular}
\end{table}

\subsection{Transfer Learning Experiment on Medical Image Datasets}

One of the most useful applications of general vision backbone models is transfer learning, where ViT models have demonstrated remarkable efficacy \cite{Zhou2021ConvNetsTransferable}. In transfer learning, a general model, trained on a large database, is finetuned on a smaller dataset, often from a different domain. In order to evaluate the transfer learning capability of ACC-ViT, we selected medical image datasets for the diversity and associated challenges \cite{Panayides2020AIDirections}. 3 different medical image datasets of different modalities and size categories were selected, as a means to critically assess how transferable the visual representations are under different configurations. The images from HAM10000 \cite{Tschandl2018TheLesions} (skin melanoma, 10,015 images), EyePACS \cite{Cuadros2009EyePACS:Screening} (diabetic retinopathy, 5,000 images) and BUSI \cite{Al-Dhabyani2020DatasetImagesb} (breast ultrasound, 1,578 images) datasets were split into 60:20:20 train-validation-test splits randomly, in a stratified manner. The tiny versions of the baseline models and ACC-ViT, pretrained on ImageNet-1K was finetuned for 100 epochs and the macro precision, recall, F1 scores along with the accuracy on the test data were computed (Table \ref{tbl:res_imgnet}). It can be observed that ACC-ViT outperformed the other models in most of the metrics, which was more noticeable on the larger datasets. For the small dataset, BUSI, ConvNeXt turned out to be the best model, ACC-ViT becoming the second best. This is expected as it has been shown in medical imaging studies that convolutional networks perform better than transformers on smaller datasets \cite{Ibtehaz2023ACC-UNet:Forthe2020s}. Out of all the models, MaxViT seemed to perform the worst, particularly for the rarer classes, which probably implies that the small-scale dataset was not sufficient to tune the model's parameters sufficiently, (please refer to the appendix for class-specific metrics).

\begin{figure*}[h]
    \centering
    \includegraphics[width=0.95\textwidth]{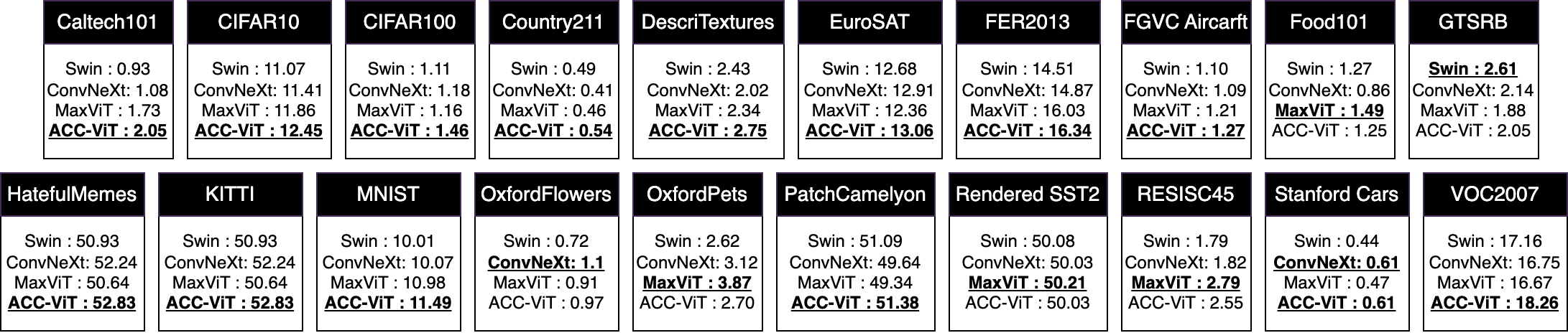}
    \caption{Zero-shot performance assessment of the tiny size models on language-vision contrastive learning.}
    \label{fig:zero}
\end{figure*}

\subsection{Frozen Backbone Feature Extraction performance assessment on Object Detection}

Frozen, pretrained vision transformers are increasingly being used as feature extractors, going beyond the conventional linear probing strategy due to the inability to capture non-linear interactions \cite{Chen2022AdaptFormer:Recognition}. The compatibility of frozen ViT backbones with additional trainable neural networks on top \cite{Alayrac2022Flamingo:Learning} or even supplementary learnable parameters in the input space \cite{Jia2022VisualTuning} has led to the development of novel methods and problem-solving protocols. In order to assess the competence of ACC-ViT as a feature extractor, we analyzed the task of object detection from a different perspective. Instead of fine-tuning the entire network \cite{Tu2022MaxViT:Transformer,Yang2022MOAT:Models}, we kept the ACC-ViT network frozen, and used Feature Pyramid Network (FPN) \cite{Lin2017FeatureDetection} and Mask R-CNN \cite{He2017MaskR-CNN} models to perform object detection and instance segmentation tasks using the default settings of the open-source MMDetection library \cite{Chen2019MMDetection:Benchmark}. The input images were resized to $224 \times 224$ and $448 \times 448$ resolutions to assess the performance at different scales and other than the default $50\%$ random flipping no other augmentations were performed.

We performed an identical experiment with the tiny versions of MaxViT and ACC-ViT and the results are presented in Table \ref{fig:mmdet}. It can be observed that both models performed better with larger resolution images and ACC-ViT-tiny achieved a better score than MaxViT-tiny in all the metrics. Therefore, ACC-ViT has the potential to be used as a frozen backbone. It should be noted that these results are not on par with the state-of-the-art object detection results, since the backbones were frozen and only the top was trained. It has been demonstrated time after time that finetuning substantially outperforms the performance of linear probing \cite{He2022MaskedLearners}. In addition, the state-of-the-art detectors use much complex detection top and training strategy with augmentation and higher resolution images \cite{Yang2022FocalNetworks}.

% \usepackage{multirow}

\subsection{Zero-shot Performance}

Supervised training or even transfer learning is infeasible for some applications due to the lack of sufficient samples from rare classes, as well as complete unavailability for the case of unseen classes \cite{Kim2023VisionLearning}. Language-image contrastive learning has become an important zero-shot task with access to large language models \cite{Alayrac2022Flamingo:Learning}. In order to assess the zero-shot performance of ACC-ViT, we used the Elevater benchmark \cite{Li2022ELEVATER:Models}. In this benchmark 20 different datasets are considered and a CLIP \cite{Radford2021LearningSupervision} model is used to assess language-augmented visual task-level transfer. We ran three random experiments with the tiny models following the default settings, and the results are presented in Fig. \ref{fig:zero}, it can be observed that ACC-ViT-tiny outperformed the other models on 13 out of 20 datasets, under this zero-shot language-image contrastive learning setting. It should be noted that these results were obtained without any additional pretraining or finetuning of the vision-language models. The models being trained on ImageNet-1K, instead of ImageNet-21K, has limited the absolute score, and so we focused on the relative scores among the models.

\subsection{Evaluation of Scaled Down Models}

In parallel to scaling up Vision Transformers to billions of parameters \cite{Dehghani2023ScalingParameters}, there have been attempts to downsize the ViT models for mobile and edge device applications \cite{Mehta2021MobileViT:Transformer}. In order to assess the performance of scaled-down versions of ACC-ViT, we simply reduced the number of layers and channels in the ACC-ViT blocks, similar to MOAT \cite{Yang2022MOAT:Models}. The performance of small-scale ACC-ViT models were compared against Mobile-Former \cite{Chen2021Mobile-Former:Transformer}, EfficientViT \cite{Liu2023EfficientViT:Attention}, MobileViT v2 \cite{Mehta2022SeparableTransformers} and tiny-MOAT family \cite{Yang2022MOAT:Models} on the ImageNet-1K dataset (Table \ref{tbl:smol}). It can be observed that in every parameter range ACC-ViT models are competitive using fewer parameters. Although tiny-MOAT-3 has a $0.36\%$ higher accuracy than ACC-ViT nano model, it has $16.77\%$ more parameters, which still puts our model on the pareto-front of Fig. \ref{fig:pareto}c and \ref{fig:pareto}d. Additionally, we had implemented a similarly sized MaxViT nano model and evaluated it on the ImageNet-1K dataset, which seemed to perform worse than ACC-ViT nano model. It should be noted that the performance of MobileViT or EfficientViT models dips a bit due to their approximate and efficient formulations of the various network operations, hence, we are not a direct competitor to them. Nevertheless, ACC-ViT outperforms the scaled-down versions of the other state-of-the-art full-fledged ViT models, such as MaxViT and MOAT.

\section{Analysis}
\label{sec:analy}

\begin{table}[b]
\centering
\caption{Ablation Study}
\label{tbl:ablation}
\begin{tabular}{lccc}
\hline
      & \#params(M) & FLOPs(G) & Acc \\ \hline
\begin{tabular}[c]{@{}l@{}}Atrous Attention\\ with MBConv\end{tabular}  &   16.907          &   4.862        & 79.498    \\ \hline
\begin{tabular}[c]{@{}l@{}}Inroducing Atrous\\ Convolution\end{tabular} &         19.33    &  4.939         & 80.017    \\ \hline
\begin{tabular}[c]
{@{}l@{}}Shared MLP\\ across attentions\end{tabular} &         14.585      & 3.473     & 81.523     \\ \hline
 
\begin{tabular}[c]{@{}l@{}}Adaptive Gating in \\ Parallel branches\end{tabular} & 16.649 & 3.812 & 82.412 \\ \hline
\begin{tabular}[c]{@{}l@{}}Replacing MLP \\ with ConvNext\end{tabular}          & 17.025 & 3.844 & 81.850 \\ \hline
%ACC-ViT-nano                                                            & 16.649      & 3.812     &           \\ \hline
\end{tabular}
\end{table}

\subsection{Ablation Study}
\label{sec:abla}
Table \ref{tbl:ablation} presents an ablation study focusing on the contributions of the different design choices in our development roadmap. We conducted our design discovery and ablation study on the nano variant, i.e., $\sim 17$ M parameter model.

We started with Atrous Attention in conjunction with MBConv, which resulted in $79.5 \%$ accuracy. Later, we used parallel atrous convolutions, increasing the accuracy only by $0.5\%$ with a considerable increase in parameters. Sharing the MLP layer across attentions, turned out quite beneficial, substantially reducing the FLOPs and increasing the accuracy. Up to this moment, the model, merely averaging the different regional information, would apparently learn the parameters quickly and reach a plateau. Implementing the gating function improved this scenario, as the model learned to focus on the key regional information dynamically, based on the input. We also attempted to replace the MLP layer with ConvNext layer, similar to MOAT, but that did not yield a satisfactory outcome and thus was discarded. %Finally, we trained the model with different kinds of augmentations and developed ACC-ViT nano model with \textcolor{red}{x} accuracy.

\subsection{Model Interpretation}
\begin{figure}[h]
    \centering
    \includegraphics[width=\columnwidth]{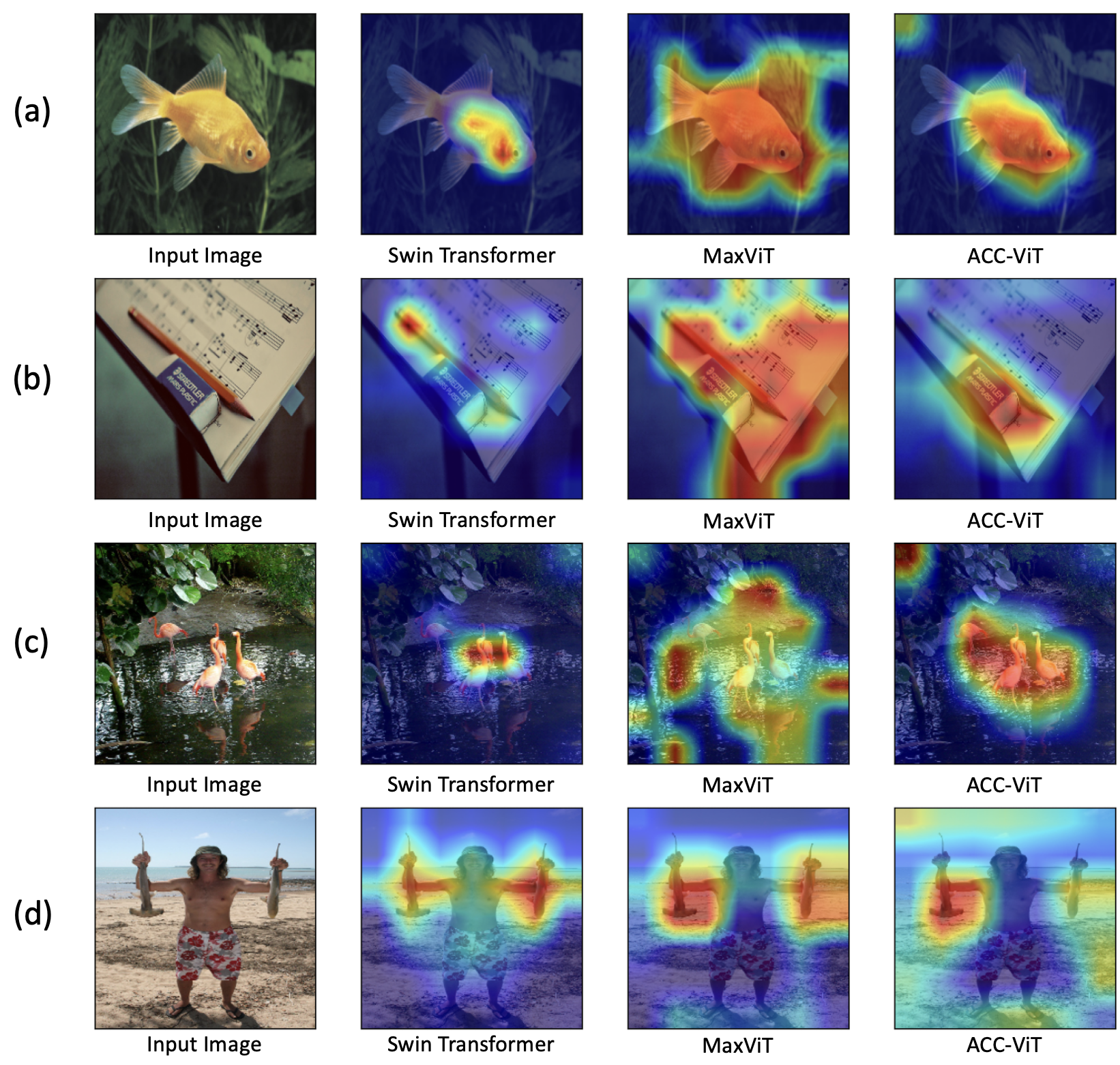}
    \caption{Model Interpretation using Grad-CAM.}
    \label{fig:grad-cam}
\end{figure}

In order to interpret and analyze what the model learns, we applied Grad-CAM \cite{Selvaraju2016Grad-CAM:Localization} on Swin, MaxViT and ACC-ViT (Fig. \ref{fig:grad-cam}). The class activation maps revealed that MaxViT tends to focus on irrelevant portions of the image, probably due to grid attention. On the contrary, Swin  seemed to often focus on a small local region. ACC-ViT apparently managed to inspect and distinguish the entire goldfish and eraser of (Fig. \ref{fig:grad-cam}a and b). Moreover, when classifying the flamingo from the example of Fig. \ref{fig:grad-cam}c, ACC-ViT focused on the entire flock of flamingos, whereas Swin and MaxViT focused on a subset and irrelevant pixels, respectively. Interestingly, when classifying  Fig. \ref{fig:grad-cam}d as hammerhead, ACC-ViT only put focus on the hammerhead fish, whereas the other transformers put emphasis on both the fishes.

\section{Conclusion}
\label{sec:conclu}

In this work, we have developed a novel attention mechanism for vision transformers, by fusing the concepts of regional and sparse attention. Seeking inspiration from atrous convolution, we designed a sparse regional attention mechanism, named Atrous Attention. Our proposed hybrid ViT architecture, ACC-ViT, maintains a balance between local and global, sparse and regional information throughout. ACC-ViT outperforms state-of-the-art vision transformer models MaxViT and MOAT. In addition, the visual representation learned by ACC-ViT seems to be quite versatile, as observed from different assessment criteria, including, finetuning, linear probing, and zero-shot learning.

The limitation of our work has been the inability to experiment with ACC-ViT to its fullest potential. Due to computational constraints, we were unable to pretrain the model on ImageNet-21K, develop extra large-scale models, process higher-resolution images. We have made our best effort to demonstrate the versatility of ACC-ViT across various tasks under a  computational budget, and we hope our proposed ideas attract sufficient attention from the computer vision community to truly assess the capabilities of the proposed transformer architecture and attention mechanism.

\clearpage
{
    \small
    \bibliographystyle{ieeenat_fullname}
    \bibliography{references}
}


\end{document}